\documentclass[10pt,twocolumn,letterpaper]{article}

\usepackage{iccv}
\usepackage{times}
\usepackage{epsfig}
\usepackage{graphicx}
\usepackage{amsmath}
\usepackage{amssymb}
\usepackage{makecell}
\usepackage{multirow}

\usepackage[pagebackref=true,breaklinks=true,letterpaper=true,colorlinks,bookmarks=false]{hyperref}

\iccvfinalcopy 

\begin{document}

\title{Swin Transformer: Hierarchical Vision Transformer using Shifted Windows}

\author{
Ze~Liu\textsuperscript{\dag\thanks{Equal contribution. \textsuperscript{\dag}Interns at MSRA. \textsuperscript{\ddag}Contact person.}}
\quad Yutong~Lin\textsuperscript{\dag*} \quad Yue~Cao\textsuperscript{*}
\quad Han~Hu\textsuperscript{*\ddag}
\quad Yixuan~Wei\textsuperscript{\dag} \\
\quad Zheng~Zhang
\quad Stephen~Lin
\quad Baining~Guo\\
{Microsoft Research Asia} \\
\small{\texttt{\{v-zeliu1,v-yutlin,yuecao,hanhu,v-yixwe,zhez,stevelin,bainguo\}@microsoft.com}}
}

\maketitle

\ificcvfinal\thispagestyle{empty}\fi

\begin{abstract}
This paper presents a new vision Transformer, called Swin Transformer, that capably serves as a general-purpose backbone for computer vision. Challenges in adapting Transformer from language to vision arise from differences between the two domains, such as large variations in the scale of visual entities and the high resolution of pixels in images compared to words in text. To address these differences, we propose a hierarchical Transformer whose representation is computed with \textbf{S}hifted \textbf{win}dows. The shifted windowing scheme brings greater efficiency by limiting self-attention computation to non-overlapping local windows while also allowing for cross-window connection. This hierarchical architecture has the flexibility to model at various scales and has linear computational complexity with respect to image size.
These qualities of Swin Transformer make it compatible with a broad range of vision tasks, including image classification (87.3 top-1 accuracy on ImageNet-1K) and dense prediction tasks such as object detection (58.7 box AP and 51.1 mask AP on COCO test-dev) and semantic segmentation (53.5 mIoU on ADE20K val). Its performance surpasses the previous state-of-the-art by a large margin of +2.7 box AP and +2.6 mask AP on COCO, and +3.2 mIoU on ADE20K, demonstrating the potential of Transformer-based models as vision backbones. The hierarchical design and the shifted window approach also prove beneficial for all-MLP architectures. The code and models are publicly available at~\url{https://github.com/microsoft/Swin-Transformer}.
\end{abstract}

\section{Introduction}

\begin{figure}
    \centering
    \includegraphics[width=1.\linewidth]{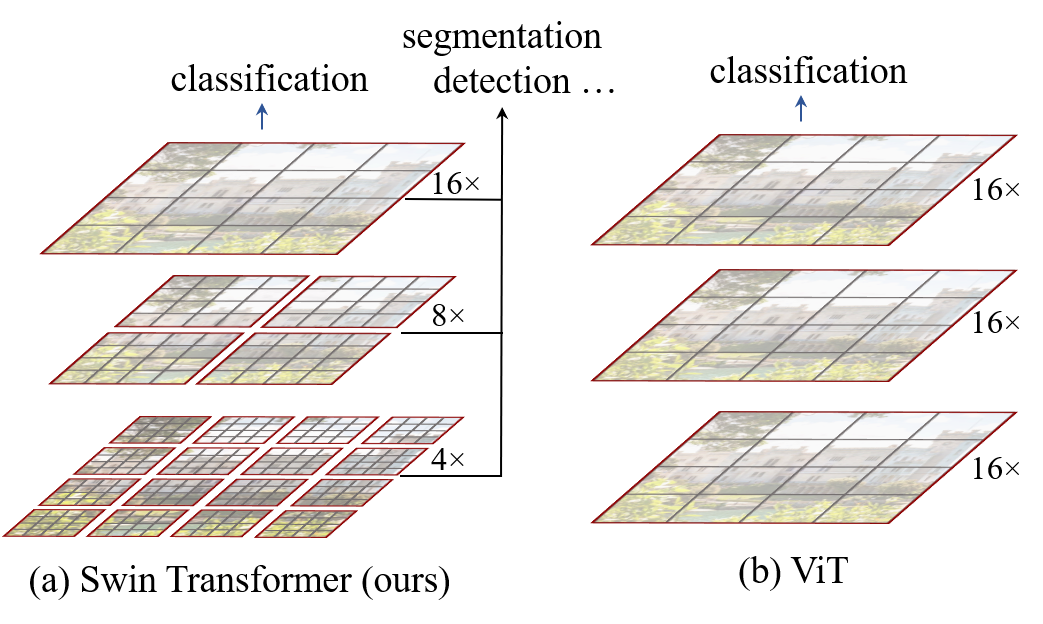}
    \caption{(a) The proposed Swin Transformer builds hierarchical feature maps by merging image patches (shown in gray) in deeper layers and has linear computation complexity to input image size due to computation of self-attention only within each local window (shown in red). It can thus serve as a general-purpose backbone for both image classification and dense recognition tasks. (b) In contrast, previous vision Transformers~\cite{dosovitskiy2020vit} produce feature maps of a single low resolution and have quadratic computation complexity to input image size due to computation of self-attention globally.}
    \label{fig:teaser}
    \vspace{-1em}
\end{figure}

Modeling in computer vision has long been dominated by convolutional neural networks (CNNs). Beginning with AlexNet~\cite{krizhevsky2012alexnet} and its revolutionary performance on the ImageNet image classification challenge, CNN architectures have evolved to become increasingly powerful through greater scale~\cite{he2015resnet,zagoruyko2016wide}, more extensive connections~\cite{huang2017densely}, and more sophisticated forms of convolution~\cite{xie2017resnext,dai2017dcnv1,zhu2018dcnv2}. With CNNs serving as backbone networks for a variety of vision tasks, these architectural advances have led to performance improvements that have broadly lifted the entire field.

On the other hand, the evolution of network architectures in natural language processing (NLP) has taken a different path, where the prevalent architecture today is instead the Transformer~\cite{vaswani2017attention}. Designed for sequence modeling and transduction tasks, the Transformer is notable for its use of attention to model long-range dependencies in the data. Its tremendous success in the language domain has led researchers to investigate its adaptation to computer vision, where it has recently demonstrated promising results on certain tasks, specifically image classification~\cite{dosovitskiy2020vit} and joint vision-language modeling~\cite{radford2021clip}.

In this paper, we seek to expand the applicability of Transformer such that it can serve as a general-purpose backbone for computer vision, as it does for NLP and as CNNs do in vision. We observe that significant challenges in transferring its high performance in the language domain to the visual domain can be explained by differences between the two modalities. One of these differences involves scale. Unlike the word tokens that serve as the basic elements of processing in language Transformers, visual elements can vary substantially in scale, a problem that receives attention in tasks such as object detection~\cite{he2017fpn,singh2018snip,singh2018sniper}. In existing Transformer-based models~\cite{vaswani2017attention,dosovitskiy2020vit}, tokens are all of a fixed scale, a property unsuitable for these vision applications. 
Another difference is the much higher resolution of pixels in images compared to words in passages of text. There exist many vision tasks such as semantic segmentation that require dense prediction at the pixel level, and this would be intractable for Transformer on high-resolution images, as the computational complexity of its self-attention is quadratic to image size. 
To overcome these issues, we propose a general-purpose Transformer backbone, called Swin Transformer, which constructs hierarchical feature maps and has linear computational complexity to image size. As illustrated in Figure~\ref{fig:teaser}(a), Swin Transformer constructs a hierarchical representation by starting from small-sized patches (outlined in gray) and gradually merging neighboring patches in deeper Transformer layers. With these hierarchical feature maps, the Swin Transformer model can conveniently leverage advanced techniques for dense prediction such as feature pyramid networks (FPN)~\cite{he2017fpn} or U-Net~\cite{ronneberger2015unet}. The linear computational complexity is achieved by computing self-attention locally within non-overlapping windows that partition an image (outlined in red). The number of patches in each window is fixed, and thus the complexity becomes linear to image size. These merits make Swin Transformer suitable as a general-purpose backbone for various vision tasks, in contrast to previous Transformer based architectures~\cite{dosovitskiy2020vit} which produce feature maps of a single resolution and have quadratic complexity.

\begin{figure}
    \centering
    \includegraphics[width=1.\linewidth]{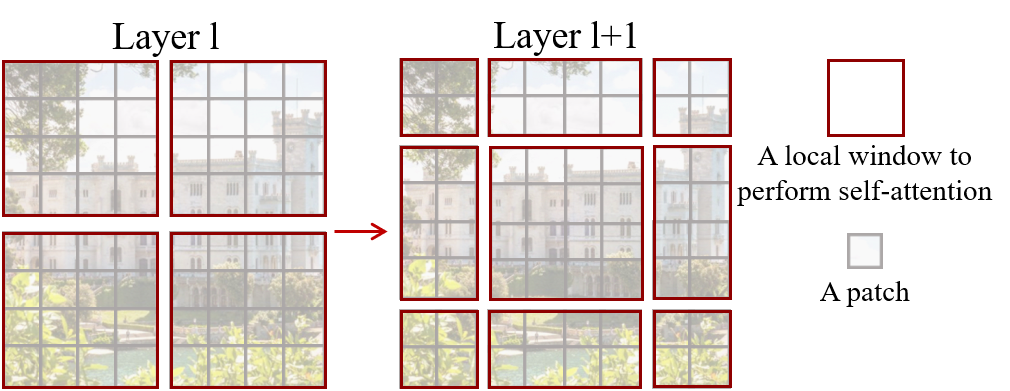}
    \caption{An illustration of the \emph{shifted window} approach for computing self-attention in the proposed Swin Transformer architecture. In layer $l$ (left), a regular window partitioning scheme is adopted, and self-attention is computed within each window. 
    In the next layer $l+1$ (right), the window partitioning is shifted, resulting in new windows. The self-attention computation in the new windows crosses the boundaries of the previous windows in layer $l$, providing connections among them.}
    \label{fig:shift_window}
    \vspace{-1em}
\end{figure}

A key design element of Swin Transformer is its \emph{shift} of the window partition between consecutive self-attention layers, as illustrated in Figure~\ref{fig:shift_window}. The shifted windows bridge the windows of the preceding layer, providing connections among them that significantly enhance modeling power (see Table~\ref{exp:ablation}). This strategy is also efficient in regards to real-world latency: all \emph{query} patches within a window share the same \emph{key} set\footnote{The \emph{query} and \emph{key} are projection vectors in a self-attention layer.}, which facilitates memory access in hardware. In contrast, earlier \emph{sliding window} based self-attention approaches~\cite{hu2019localrelation,ramachandran2019stand} suffer from low latency on general hardware due to different \emph{key} sets for different \emph{query} pixels\footnote{While there are efficient methods to implement a sliding-window based convolution layer on general hardware, thanks to its shared kernel weights across a feature map, it is difficult for a sliding-window based self-attention layer to have efficient memory access in practice.}. Our experiments show that the proposed \emph{shifted window} approach has much lower latency than the \emph{sliding window} method, yet is similar in modeling power (see Tables~\ref{tab:ablation-selfatt-efficient} and \ref{exp:abaltion-selfatt-acc}). The shifted window approach also proves beneficial for all-MLP architectures~\cite{tolstikhin2021mlpmixer}.

The proposed Swin Transformer achieves strong performance on the recognition tasks of image classification, object detection and semantic segmentation. It outperforms the ViT / DeiT~\cite{dosovitskiy2020vit,touvron2020deit} and ResNe(X)t models~\cite{he2015resnet,xie2017resnext} significantly with similar latency on the three tasks. Its 58.7 box AP and 51.1 mask AP on the COCO test-dev set surpass the previous state-of-the-art results by +2.7 box AP (Copy-paste~\cite{ghiasi2020copy} without external data) and +2.6 mask AP (DetectoRS~\cite{qiao2020detectors}). On ADE20K semantic segmentation, it obtains 53.5 mIoU on the val set, an improvement of +3.2 mIoU over the previous state-of-the-art (SETR~\cite{zheng2020SETR}). It also achieves a top-1 accuracy of 87.3\% on ImageNet-1K image classification.

It is our belief that a unified architecture across computer vision and natural language processing could benefit both fields, since it would facilitate joint modeling of visual and textual signals and the modeling knowledge from both domains can be more deeply shared. We hope that Swin Transformer's strong performance on various vision problems can drive this belief deeper in the community and encourage unified modeling of vision and language signals.

\section{Related Work}

\paragraph{CNN and variants} CNNs serve as the standard network model throughout computer vision. While the CNN has existed for several decades~\cite{lecun1998lenet}, it was not until the introduction of AlexNet~\cite{krizhevsky2012alexnet} that the CNN took off and became mainstream. Since then, deeper and more effective convolutional neural architectures have been proposed to further propel the deep learning wave in computer vision, e.g., VGG~\cite{simonyan2014vgg}, GoogleNet~\cite{szegedy2015googlenet}, ResNet~\cite{he2015resnet}, DenseNet~\cite{huang2017densely}, HRNet~\cite{wang2020deep}, and EfficientNet~\cite{tan2019efficientnet}. In addition to these architectural advances, there has also been much work on improving individual convolution layers, such as depth-wise convolution~\cite{xie2017resnext} and deformable convolution~\cite{dai2017dcnv1,zhu2018dcnv2}. While the CNN and its variants are still the primary backbone architectures for computer vision applications, we highlight the strong potential of Transformer-like architectures for unified modeling between vision and language. Our work achieves strong performance on several basic visual recognition tasks, and we hope it will contribute to a modeling shift. 

\paragraph{Self-attention based backbone architectures} Also inspired by the success of self-attention layers and Transformer architectures in the NLP field, some works employ self-attention layers to replace some or all of the spatial convolution layers in the popular ResNet~\cite{hu2019localrelation,ramachandran2019stand,zhao2020SAN}. In these works, the self-attention is computed within a local window of each pixel to expedite optimization~\cite{hu2019localrelation}, and they achieve slightly better accuracy/FLOPs trade-offs than the counterpart ResNet architecture. However, their costly memory access causes their actual latency to be significantly larger than that of the convolutional networks~\cite{hu2019localrelation}. Instead of using sliding windows, we propose to \emph{shift} windows between consecutive layers, which allows for a more efficient implementation in general hardware.

\paragraph{Self-attention/Transformers to complement CNNs} Another line of work is to augment a standard CNN architecture with self-attention layers or Transformers. The self-attention layers can complement backbones \cite{wang2017non,cao2019gcnet,bello2020attention,yin2020DNL,fu2019danet,yuan2018ocnet,srinivas2021bottlenecktrans} or head networks~\cite{hu2018relation,gu2018learnregionfeat} by providing the capability to encode distant dependencies or heterogeneous interactions. More recently, the encoder-decoder design in Transformer has been applied for the object detection and instance segmentation tasks~\cite{carion2020detr,chi2020relationnet++,zhu2020deformabledetr,sun2020sparsercnn}. Our work explores the adaptation of Transformers for basic visual feature extraction and is complementary to these works.

\paragraph{Transformer based vision backbones} Most related to our work is the Vision Transformer (ViT)~\cite{dosovitskiy2020vit} and its follow-ups~\cite{touvron2020deit,yuan2021t2t,chu2021postrans,han2021tnt,wang2021pvt}. The pioneering work of ViT directly applies a Transformer architecture on non-overlapping medium-sized image patches for image classification. It achieves an impressive speed-accuracy trade-off on image classification compared to convolutional networks. While ViT requires large-scale training datasets (i.e., JFT-300M) to perform well, DeiT~\cite{touvron2020deit} introduces several training strategies that allow ViT to also be effective using the smaller ImageNet-1K dataset. The results of ViT on image classification are encouraging, but its architecture is unsuitable for use as a general-purpose backbone network on dense vision tasks or when the input image resolution is high, due to its low-resolution feature maps and the quadratic increase in complexity with image size. There are a few works applying ViT models to the dense vision tasks of object detection and semantic segmentation by direct upsampling or deconvolution but with relatively lower performance~\cite{beal2020toward,zheng2020SETR}. Concurrent to our work are some that modify the ViT architecture~\cite{yuan2021t2t,chu2021postrans,han2021tnt} for better image classification. Empirically, we find our Swin Transformer architecture to achieve the best speed-accuracy trade-off among these methods on image classification, even though our work focuses on general-purpose performance rather than specifically on classification. Another concurrent work~\cite{wang2021pvt} explores a similar line of thinking to build multi-resolution feature maps on Transformers. Its complexity is still quadratic to image size, while ours is linear and also operates locally which has proven beneficial in modeling the high correlation in visual signals~\cite{hubel1962receptive,fukushima1975cognitron,lecun1999object}. Our approach is both efficient and effective, achieving state-of-the-art accuracy on both COCO object detection and ADE20K semantic segmentation. 

\section{Method}

\begin{figure*}
  \centering
  \includegraphics[width=1.02\linewidth]{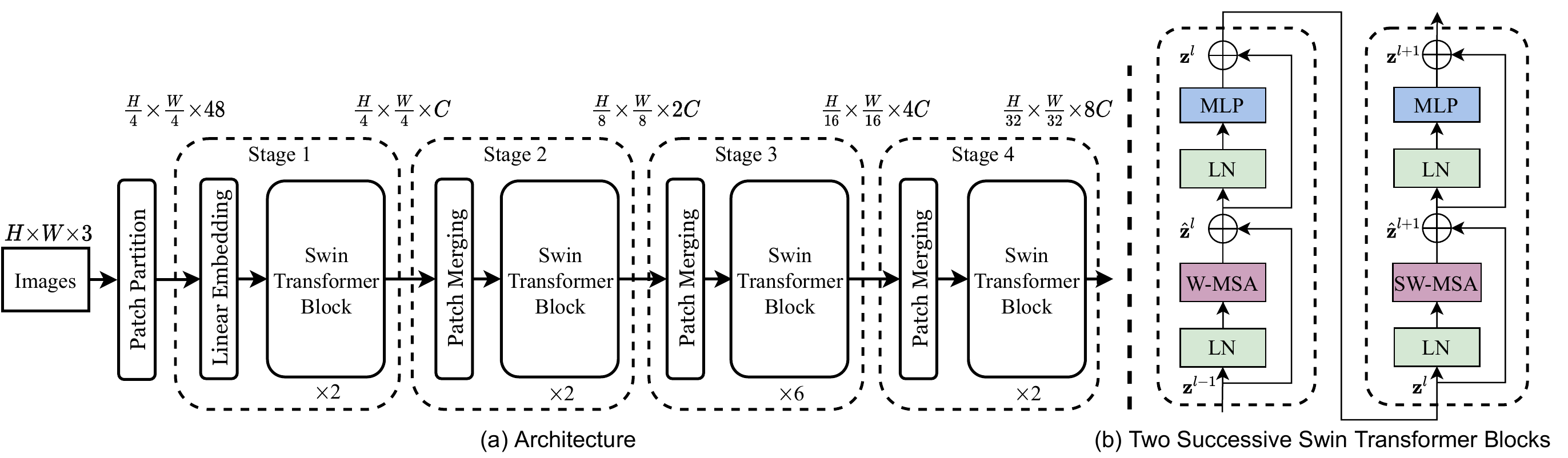}
  \caption{(a) The architecture of a Swin Transformer (Swin-T); (b) two successive Swin Transformer Blocks (notation presented with Eq.~(\ref{eq.swin})). W-MSA and SW-MSA are multi-head self attention modules with regular and shifted windowing configurations, respectively. }
  \label{fig:arch}
\end{figure*}

\subsection{Overall Architecture}

An overview of the Swin Transformer architecture is presented in Figure~\ref{fig:arch}, which illustrates the tiny version (Swin-T). It first
splits an input RGB image into non-overlapping patches by a patch splitting module, like ViT. Each patch is treated as a ``token'' and its feature is set as a concatenation of the raw pixel RGB values. In our implementation, we use a patch size of $4\times 4$ and thus the feature dimension of each patch is $4\times4\times 3=48$. A linear embedding layer is applied on this raw-valued feature to project it to an arbitrary dimension (denoted as $C$).

Several Transformer blocks with modified self-attention computation ({\em Swin Transformer blocks}) are applied on these patch tokens. The Transformer blocks maintain the number of tokens ($\frac{H}{4} \times \frac{W}{4}$), and together with the linear embedding are referred to as ``Stage 1''.

To produce a hierarchical representation, the number of tokens is reduced by patch merging layers as the network gets deeper. The first patch merging layer concatenates the features of each group of $2\times 2$ neighboring patches, and applies a linear layer on the $4C$-dimensional concatenated features. This reduces the number of tokens by a multiple of $2\times2=4$ ($2\times$ downsampling of resolution), and the output dimension is set to $2C$. Swin Transformer blocks are applied afterwards for feature transformation, with the resolution kept at $\frac{H}{8} \times \frac{W}{8}$. This first block of patch merging and feature transformation is denoted as ``Stage 2''. The procedure is repeated twice, as ``Stage 3'' and ``Stage 4'', with output resolutions of $\frac{H}{16} \times \frac{W}{16}$ and $\frac{H}{32} \times \frac{W}{32}$, respectively. These stages jointly produce a hierarchical representation, with the same feature map resolutions as those of typical convolutional networks, e.g., VGG~\cite{simonyan2014vgg} and ResNet~\cite{he2015resnet}. As a result, the proposed architecture can conveniently replace the backbone networks in existing methods for various vision tasks.

\paragraph{Swin Transformer block} Swin Transformer is built by replacing the standard multi-head self attention (MSA) module in a Transformer block by a module based on shifted windows (described in Section~\ref{sec:shifted}), with other layers kept the same. As illustrated in Figure~\ref{fig:arch}(b), a Swin Transformer block consists of a shifted window based MSA module, followed by a 2-layer MLP with GELU non-linearity in between. A LayerNorm (LN) layer is applied before each MSA module and each MLP, and a residual connection is applied after each module.

\subsection{Shifted Window based Self-Attention}
\label{sec:shifted}

The standard Transformer architecture~\cite{vaswani2017attention} and its adaptation for image classification~\cite{dosovitskiy2020vit} both conduct global self-attention, where the relationships between a token and all other tokens are computed. The global computation leads to quadratic complexity with respect to the number of tokens, making it unsuitable for many vision problems requiring an immense set of tokens for dense prediction or to represent a high-resolution image.

\paragraph{Self-attention in non-overlapped windows}
For efficient modeling, we propose to compute self-attention within local windows. The windows are arranged to evenly partition the image in a non-overlapping manner. Supposing each window contains $M\times M$ patches, the computational complexity of a global MSA module and a window based one on an image of $h\times w$ patches are\footnote{We omit SoftMax computation in determining complexity.}:
\begin{align}
&\Omega (\text{MSA}) = 4hwC^2 + 2 (hw)^2C, \\
&\Omega (\text{W-MSA}) = 4hwC^2 + 2 M^2 hwC,
\end{align}
where the former is quadratic to patch number $hw$, and the latter is linear when $M$ is fixed (set to $7$ by default). Global self-attention computation is generally unaffordable for a large $hw$, while the window based self-attention is scalable.

\paragraph{Shifted window partitioning in successive blocks} The window-based self-attention module lacks connections across windows, which limits its modeling power. To introduce cross-window connections while maintaining the efficient computation of non-overlapping windows, we propose a shifted window partitioning approach which alternates between two partitioning configurations in consecutive Swin Transformer blocks.

As illustrated in Figure~\ref{fig:shift_window}, the first module uses a regular window partitioning strategy which starts from the top-left pixel, and the $8\times 8$ feature map is evenly partitioned into $2\times2$ windows of size $4\times4$ ($M=4$). Then, the next module adopts a windowing configuration that is shifted from that of the preceding layer, by displacing the windows by $ (\lfloor\frac{M}{2} \rfloor, \lfloor\frac{M}{2} \rfloor)$ pixels from the regularly partitioned windows. 

With the shifted window partitioning approach, consecutive Swin Transformer blocks are computed as
\begin{align}
    &{{\hat{\bf{z}}}^{l}} = \text{W-MSA}\left( {\text{LN}\left( {{{\bf{z}}^{l - 1}}} \right)} \right) + {\bf{z}}^{l - 1},\nonumber\\
    &{{\bf{z}}^l} = \text{MLP}\left( {\text{LN}\left( {{{\hat{\bf{z}}}^{l}}} \right)} \right) + {{\hat{\bf{z}}}^{l}},\nonumber\\
    &{{\hat{\bf{z}}}^{l+1}} = \text{SW-MSA}\left( {\text{LN}\left( {{{\bf{z}}^{l}}} \right)} \right) + {\bf{z}}^{l}, \nonumber\\
    &{{\bf{z}}^{l+1}} = \text{MLP}\left( {\text{LN}\left( {{{\hat{\bf{z}}}^{l+1}}} \right)} \right) + {{\hat{\bf{z}}}^{l+1}}, \label{eq.swin}
\end{align}
where ${\hat{\bf{z}}}^l$ and ${\bf{z}}^l$ denote the output features of the (S)W-MSA module and the MLP module for block $l$, respectively; $\text{W-MSA}$ and $\text{SW-MSA}$ denote window based multi-head self-attention using regular and shifted window partitioning configurations, respectively.

The shifted window partitioning approach introduces connections between neighboring non-overlapping windows in the previous layer and is found to be effective in image classification, object detection, and semantic segmentation, as shown in Table~\ref{exp:ablation}.

\begin{figure}
    \centering
    \includegraphics[width=1.\linewidth]{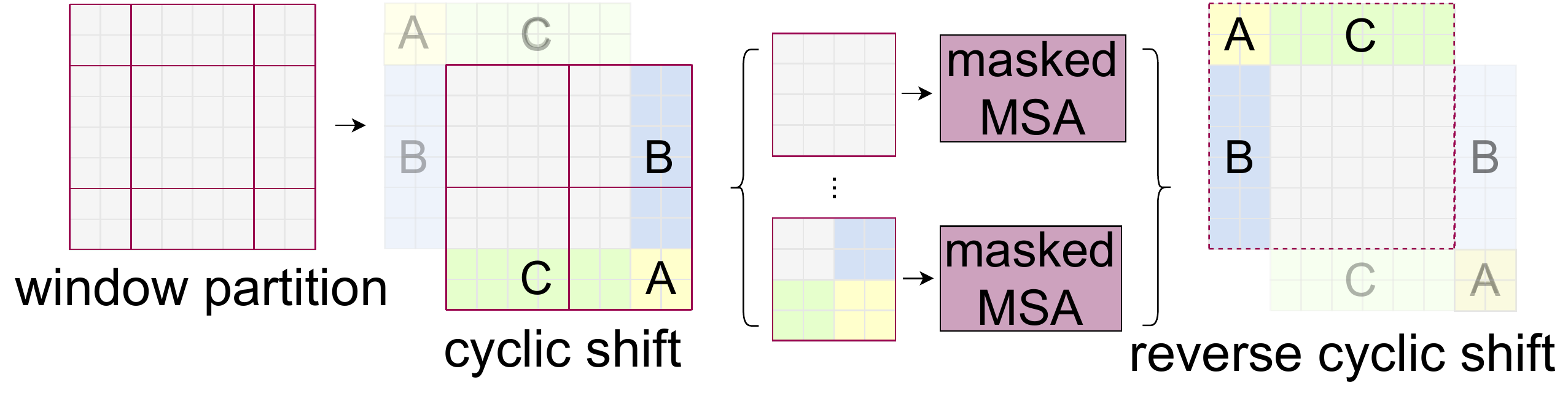}
    \caption{Illustration of an efficient batch computation approach for self-attention in shifted window partitioning. }
    \label{fig:shifted_window_att}
\end{figure}

\paragraph{Efficient batch computation for shifted configuration} An issue with shifted window partitioning is that it will result in more windows, from $\lceil \frac{h}{M}\rceil  \times \lceil\frac{w}{M}\rceil$  to $(\lceil \frac{h}{M}\rceil +1) \times (\lceil\frac{w}{M}\rceil + 1)$ in the shifted configuration, and some of the windows will be smaller than $M\times M$\footnote{To make the window size $(M, M)$ divisible by the feature map size of $(h, w)$, bottom-right padding is employed on the feature map if needed.}. A naive solution is to pad the smaller windows to a size of $M\times M$ and mask out the padded values when computing attention. When the number of windows in regular partitioning is small, e.g. $2\times 2$, the increased computation with this naive solution is considerable ($2\times 2 \rightarrow 3\times 3$, which is 2.25 times greater). Here, we propose a \emph{more efficient batch computation approach} by cyclic-shifting toward the top-left direction, as illustrated in Figure~\ref{fig:shifted_window_att}. After this shift, a batched window may be composed of several sub-windows that are not adjacent in the feature map, so a masking mechanism is employed to limit self-attention computation to within each sub-window.
With the cyclic-shift, the number of batched windows remains the same as that of regular window partitioning, and thus is also efficient. The low latency of this approach is shown in Table~\ref{tab:ablation-selfatt-efficient}.

\paragraph{Relative position bias} In computing self-attention, we follow~\cite{raffel2019t5,bao2020unilmv2,hu2018relation,hu2019localrelation} by including a relative position bias $B \in \mathbb{R}^{M^2 \times M^2}$ to each head in computing similarity:
\begin{equation}
\label{eq.att}
    \text{Attention}(Q, K, V) = \text{SoftMax}(QK^T/\sqrt{d}+B)V,
\end{equation}
where $Q, K, V \in \mathbb{R}^{M^2\times d}$ are the \emph{query}, \emph{key} and \emph{value} matrices; $d$ is the \emph{query}/\emph{key} dimension, and $M^2$ is the number of patches in a window. Since the relative position along each axis lies in the range $[-M+1, M-1]$, we parameterize a smaller-sized bias matrix $\hat{B} \in \mathbb{R}^{(2M-1)\times (2M-1)}$, and values in $B$ are taken from $\hat{B}$.

We observe significant improvements over counterparts without this bias term or that use absolute position embedding, as shown in Table~\ref{exp:ablation}. Further adding absolute position embedding to the input as in~\cite{dosovitskiy2020vit} drops performance slightly, thus it is not adopted in our implementation.

The learnt relative position bias in pre-training can be also used to initialize a model for fine-tuning with a different window size through bi-cubic interpolation~\cite{dosovitskiy2020vit,touvron2020deit}.

\subsection{Architecture Variants}

We build our base model, called Swin-B, to have of model size and computation complexity similar to ViT-B/DeiT-B. We also introduce Swin-T, Swin-S and Swin-L, which are versions of about $0.25\times$, $0.5\times$ and $2\times$ the model size and computational complexity, respectively. Note that the complexity of Swin-T and Swin-S are similar to those of ResNet-50 (DeiT-S) and ResNet-101, respectively. The window size is set to $M=7$ by default. The query dimension of each head is $d=32$, and the expansion layer of each MLP is $\alpha=4$, for all experiments. The architecture hyper-parameters of these model variants are:
\begin{itemize}
    \item Swin-T: $C=96$, layer numbers = $\{2, 2, 6, 2\}$
    \item Swin-S: $C=96$, layer numbers =$\{2, 2, 18, 2\}$
    \item Swin-B: $C=128$, layer numbers =$\{2, 2, 18, 2\}$
    \item Swin-L: $C=192$, layer numbers =$\{2, 2, 18, 2\}$ 
\end{itemize}
where $C$ is the channel number of the hidden layers in the first stage. The model size, theoretical computational complexity (FLOPs), and throughput of the model variants for ImageNet image classification are listed in Table~\ref{exp:imagenet-system}. 

\section{Experiments}

We conduct experiments on ImageNet-1K image classification~\cite{deng2009imagenet}, COCO object detection~\cite{lin2014coco}, and ADE20K semantic segmentation~\cite{zhou2018semantic}. In the following, we first compare the proposed Swin Transformer architecture with the previous state-of-the-arts on the three tasks. Then, we ablate the important design elements of Swin Transformer.

\subsection{Image Classification on ImageNet-1K}

\paragraph{Settings} For image classification, we benchmark the proposed Swin Transformer on ImageNet-1K~\cite{deng2009imagenet}, which contains 1.28M training images and 50K validation images from 1,000 classes. The top-1 accuracy on a single crop is reported. We consider two training settings:
\begin{itemize}
    \item \emph{Regular ImageNet-1K training}. This setting mostly follows~\cite{touvron2020deit}. We employ an AdamW~\cite{kingma2014adam} optimizer for 300 epochs using a cosine decay learning rate scheduler and 20 epochs of linear warm-up. A batch size of 1024, an initial learning rate of 0.001, and a weight decay of 0.05 are used. We include most of the augmentation and regularization strategies of~\cite{touvron2020deit} in training, except for repeated augmentation~\cite{hoffer2020augment} and EMA~\cite{polyak1992acceleration}, which do not enhance performance. Note that this is contrary to~\cite{touvron2020deit} where repeated augmentation is crucial to stabilize the training of ViT. 
    \item \emph{Pre-training on ImageNet-22K and fine-tuning on ImageNet-1K}. We also pre-train on the larger ImageNet-22K dataset, which contains 14.2 million images and 22K classes. We employ an AdamW optimizer for 90 epochs using a linear decay learning rate scheduler with a 5-epoch linear warm-up. A batch size of 4096, an initial learning rate of 0.001, and a weight decay of 0.01 are used. In ImageNet-1K fine-tuning, we train the models for 30 epochs with a batch size of 1024, a constant learning rate of $10^{-5}$, and a weight decay of $10^{-8}$.
\end{itemize}

\paragraph{Results with regular ImageNet-1K training} Table~\ref{exp:imagenet-system}(a) presents comparisons to other backbones, including both Transformer-based and ConvNet-based, using regular ImageNet-1K training.

Compared to the previous state-of-the-art Transformer-based architecture, i.e. DeiT~\cite{touvron2020deit}, Swin Transformers noticeably surpass the counterpart DeiT architectures with similar complexities: +1.5\% for Swin-T (81.3\%) over DeiT-S (79.8\%) using 224$^2$ input, and +1.5\%/1.4\% for Swin-B (83.3\%/84.5\%) over DeiT-B (81.8\%/83.1\%) using 224$^2$/384$^2$ input, respectively.

Compared with the state-of-the-art ConvNets, i.e. RegNet~\cite{radosavovic2020regnet} and EfficientNet~\cite{tan2019efficientnet}, the Swin Transformer achieves a slightly better speed-accuracy trade-off. Noting that while RegNet~\cite{radosavovic2020regnet} and EfficientNet~\cite{tan2019efficientnet} are obtained via a thorough architecture search, the proposed Swin Transformer is adapted from the standard Transformer and has strong potential for further improvement.

\paragraph{Results with ImageNet-22K pre-training} We also pre-train the larger-capacity Swin-B and Swin-L on ImageNet-22K. Results fine-tuned on ImageNet-1K image classification are shown in Table~\ref{exp:imagenet-system}(b). For Swin-B, the ImageNet-22K pre-training brings 1.8\%$\sim$1.9\% gains over training on ImageNet-1K from scratch. Compared with the previous best results for ImageNet-22K pre-training, our models achieve significantly better speed-accuracy trade-offs: Swin-B obtains 86.4\% top-1 accuracy, which is 2.4\% higher than that of ViT with similar inference throughput (84.7 vs. 85.9 images/sec) and slightly lower FLOPs (47.0G vs. 55.4G). The larger Swin-L model achieves 87.3\% top-1 accuracy, +0.9\% better than that of the Swin-B model. 

\begin{table}[t]
\centering
\small
\addtolength{\tabcolsep}{-5.pt}
\begin{tabular}{c|cccc|c}
\Xhline{1.0pt}
\multicolumn{6}{c}{\textbf{(a) Regular ImageNet-1K trained models}} \\
method &  \begin{tabular}[c]{@{}c@{}}image \\ size\end{tabular} & \#param. & FLOPs & \begin{tabular}[c]{@{}c@{}}throughput\\ (image / s)\end{tabular} & \begin{tabular}[c]{@{}c@{}}ImageNet \\ top-1 acc.\end{tabular} \\
\hline
RegNetY-4G~\cite{radosavovic2020regnet} & 224$^2$ & 21M & 4.0G & 1156.7 & 80.0 \\
RegNetY-8G~\cite{radosavovic2020regnet} & 224$^2$ & 39M & 8.0G & 591.6 & 81.7 \\
RegNetY-16G~\cite{radosavovic2020regnet} & 224$^2$ & 84M & 16.0G & 334.7 & 82.9 \\
\hline
EffNet-B3~\cite{tan2019efficientnet} & 300$^2$ & 12M & 1.8G & 732.1 & 81.6 \\
EffNet-B4~\cite{tan2019efficientnet} & 380$^2$ & 19M & 4.2G & 349.4 & 82.9 \\
EffNet-B5~\cite{tan2019efficientnet} & 456$^2$ & 30M & 9.9G & 169.1 & 83.6 \\
EffNet-B6~\cite{tan2019efficientnet} & 528$^2$ & 43M & 19.0G & 96.9 & 84.0 \\
EffNet-B7~\cite{tan2019efficientnet} & 600$^2$ & 66M & 37.0G & 55.1 & 84.3 \\
\hline
ViT-B/16~\cite{dosovitskiy2020vit} & 384$^2$ & 86M & 55.4G & 85.9 & 77.9 \\
ViT-L/16~\cite{dosovitskiy2020vit} & 384$^2$ & 307M & 190.7G & 27.3 & 76.5 \\
\hline
DeiT-S~\cite{touvron2020deit} & 224$^2$ & 22M & 4.6G & 940.4 & 79.8 \\
DeiT-B~\cite{touvron2020deit} & 224$^2$ & 86M & 17.5G & 292.3 & 81.8 \\
DeiT-B~\cite{touvron2020deit} & 384$^2$ & 86M & 55.4G & 85.9 & 83.1 \\
\hline
Swin-T & 224$^2$ & 29M & 4.5G & 755.2 & 81.3 \\
Swin-S & 224$^2$ & 50M & 8.7G & 436.9 & 83.0 \\
Swin-B & 224$^2$ & 88M & 15.4G & 278.1 & 83.5 \\
Swin-B & 384$^2$ & 88M & 47.0G & 84.7 & 84.5 \\
\Xhline{1.0pt}
\multicolumn{6}{c}{\textbf{(b) ImageNet-22K pre-trained models}}  \\  
method & \begin{tabular}[c]{@{}c@{}}image \\ size\end{tabular} & \#param. & FLOPs & \begin{tabular}[c]{@{}c@{}}throughput\\ (image / s)\end{tabular} & \begin{tabular}[c]{@{}c@{}}ImageNet \\ top-1 acc.\end{tabular} \\
\hline
R-101x3~\cite{kolesnikov2019bigtransfer} & 384$^2$ & 388M & 204.6G & - & 84.4 \\
R-152x4~\cite{kolesnikov2019bigtransfer} & 480$^2$ & 937M & 840.5G & - & 85.4 \\
\hline
ViT-B/16~\cite{dosovitskiy2020vit} & 384$^2$ & 86M & 55.4G & 85.9 & 84.0 \\
ViT-L/16~\cite{dosovitskiy2020vit} & 384$^2$ & 307M & 190.7G & 27.3 & 85.2 \\
\hline
Swin-B & 224$^2$ & 88M & 15.4G & 278.1 & 85.2 \\
Swin-B & 384$^2$ & 88M & 47.0G & 84.7 & 86.4 \\
Swin-L & 384$^2$ & 197M & 103.9G & 42.1 & 87.3  \\
\Xhline{1.0pt}
\end{tabular}
\normalsize
\caption{Comparison of different backbones on ImageNet-1K classification. Throughput is measured using the GitHub repository of~\cite{rw2019timm} and a V100 GPU, following~\cite{touvron2020deit}.}
\label{exp:imagenet-system}
\end{table}

\subsection{Object Detection on COCO}

\paragraph{Settings} Object detection and instance segmentation experiments are conducted on COCO 2017, which contains 118K training, 5K validation and 20K test-dev images. An ablation study is performed using the validation set, and a system-level comparison is reported on test-dev. For the ablation study, we consider four typical object detection frameworks: 
Cascade Mask R-CNN~\cite{he2017mask,cai2018cascade}, ATSS~\cite{zhang2020bridging}, RepPoints v2~\cite{chen2020reppointsv2}, and Sparse RCNN~\cite{sun2020sparsercnn} in mmdetection~\cite{chen2019mmdetection}. For these four frameworks, we utilize the same settings: multi-scale training~\cite{carion2020detr,sun2020sparsercnn} (resizing the input such that the shorter side is between 480 and 800 while the longer side is at most 1333), AdamW~\cite{loshchilov2017decoupled} optimizer (initial learning rate of 0.0001, weight decay of 0.05, and batch size of 16), and 3x schedule (36 epochs).
For system-level comparison, we adopt an improved HTC~\cite{chen2019htc} (denoted as HTC++) with instaboost~\cite{fang2019instaboost}, stronger multi-scale training~\cite{cao2019gcnet}, 6x schedule (72 epochs), soft-NMS~\cite{Bodla2017softnms}, and ImageNet-22K pre-trained model as initialization.

We compare our Swin Transformer to standard ConvNets, i.e. ResNe(X)t, and previous Transformer networks, e.g. DeiT. The comparisons are conducted by changing only the backbones with other settings unchanged. Note that while Swin Transformer and ResNe(X)t are directly applicable to all the above frameworks because of their hierarchical feature maps, DeiT only produces a single resolution of feature maps and cannot be directly applied. For fair comparison, we follow~\cite{zheng2020SETR} to construct hierarchical feature maps for DeiT using deconvolution layers. 

\begin{table}[t]
\small
\addtolength{\tabcolsep}{-4.9pt}
\begin{tabular}{cc|ccc|ccc}
\Xhline{1.0pt}
\multicolumn{7}{c}{\textbf{(a) Various frameworks}}  \\  
Method & Backbone & AP$^\text{box}$ & AP$^\text{box}_\text{50}$ & AP$^\text{box}_\text{75}$ & \#param. & FLOPs & FPS \\
\hline
\multirow{2}{*}{\makecell{Cascade \\Mask R-CNN}} & R-50 & 46.3 & 64.3 & 50.5 & 82M & 739G & 18.0 \\
 & Swin-T & \textbf{50.5} & \textbf{69.3} & \textbf{54.9} & 86M & 745G & 15.3 \\
\hline
\multirow{2}{*}{ATSS} & R-50 & 43.5	& 61.9 & 47.0 & 32M & 205G & 28.3 \\
 & Swin-T & \textbf{47.2} & \textbf{66.5} & \textbf{51.3} & 36M & 215G & 22.3 \\
\hline
\multirow{2}{*}{RepPointsV2} & R-50 & 46.5 & 64.6 & 50.3 & 42M & 274G & 13.6 \\
 & Swin-T &\textbf{50.0} & \textbf{68.5} & \textbf{54.2} & 45M & 283G & 12.0 \\	
 \hline
\multirow{2}{*}{\makecell{Sparse\\R-CNN}} & R-50 & 44.5 & 63.4 & 48.2 & 106M & 166G & 21.0 \\
 & Swin-T & \textbf{47.9} & \textbf{67.3} & \textbf{52.3} & 110M & 172G & 18.4 \\
\Xhline{1.0pt}
\end{tabular}
\addtolength{\tabcolsep}{-0.9pt}
\begin{tabular}{c|ccc|ccc|ccc}
\multicolumn{9}{c}{\textbf{(b) Various backbones w. Cascade Mask R-CNN}}  \\  
& AP$^\text{box}$ & AP$^\text{box}_\text{50}$ & AP$^\text{box}_\text{75}$ & AP$^\text{mask}$ & AP$^\text{mask}_\text{50}$ & AP$^\text{mask}_\text{75}$ & param & FLOPs & FPS \\
\hline
DeiT-S$^\dag$ & 48.0 & 67.2 & 51.7 & 41.4 & 64.2 & 44.3 & 80M & 889G & 10.4 \\
R50 & 46.3 & 64.3 & 50.5 & 40.1 & 61.7 & 43.4 & 82M & 739G & 18.0 \\
Swin-T & \textbf{50.5} & \textbf{69.3} & \textbf{54.9} & \textbf{43.7} & \textbf{66.6} & \textbf{47.1} & 86M & 745G & 15.3 \\
\hline
X101-32 & 48.1 & 66.5 & 52.4 & 41.6 & 63.9 & 45.2 & 101M & 819G & 12.8 \\
Swin-S & \textbf{51.8} & \textbf{70.4} & \textbf{56.3} & \textbf{44.7} & \textbf{67.9} & \textbf{48.5} & 107M & 838G & 12.0 \\
\hline
X101-64 & 48.3 & 66.4 & 52.3 & 41.7 & 64.0 & 45.1 & 140M & 972G & 10.4 \\
Swin-B & \textbf{51.9} & \textbf{70.9} & \textbf{56.5} & \textbf{45.0} & \textbf{68.4} & \textbf{48.7} & 145M & 982G & 11.6\\
\Xhline{1.0pt}
\end{tabular}
\addtolength{\tabcolsep}{0.9pt}
\begin{tabular}{c|cc|cc|cc}
\multicolumn{7}{c}{\textbf{(c) System-level Comparison}}  \\  
\multirow{2}{*}{Method} & \multicolumn{2}{c|}{mini-val} & \multicolumn{2}{c|}{test-dev} & \multirow{2}{*}{\#param.} & \multirow{2}{*}{FLOPs}  \\
 & AP$^\text{box}$ & AP$^\text{mask}$ & AP$^\text{box}$ & AP$^\text{mask}$ &  &  \\
\hline
RepPointsV2*~\cite{chen2020reppointsv2} & - & - & 52.1 & - & - & - \\
GCNet*~\cite{cao2019gcnet} & 51.8 & 44.7 & 52.3 & 45.4 & - & 1041G\\
RelationNet++*~\cite{chi2020relationnet++} & - & - & 52.7 & - & - & - \\
SpineNet-190~\cite{du2020spinenet} & 52.6 & - & 52.8 & - & 164M & 1885G \\
ResNeSt-200*~\cite{zhang2020resnest} & 52.5 & - & 53.3 & 47.1 & - & - \\
EfficientDet-D7~\cite{tan2020efficientdet} & 54.4 & - & 55.1 & - & 77M & 410G \\
DetectoRS*~\cite{qiao2020detectors} & - & - & 55.7 & 48.5 & - & - \\
YOLOv4 P7*~\cite{bochkovskiy2020yolov4} & - & - & 55.8 & - & - & - \\
Copy-paste~\cite{ghiasi2020copy} & 55.9 & 47.2 & 56.0 & 47.4 & 185M & 1440G \\
\hline
X101-64 (HTC++) & 52.3 & 46.0 & - & - & 155M & 1033G \\
\hline
Swin-B (HTC++) & 56.4 & 49.1 & - & - & 160M & 1043G \\
Swin-L (HTC++) & 57.1 & 49.5 & 57.7 & 50.2 & 284M & 1470G \\
Swin-L (HTC++)* & \textbf{58.0} & \textbf{50.4} & \textbf{58.7} & \textbf{51.1} & 284M & - \\
\Xhline{1.0pt}
\end{tabular}
\caption{Results on COCO object detection and instance segmentation. $^\dag$denotes that additional decovolution layers are used to produce hierarchical feature maps. * indicates multi-scale testing.}
\label{exp:coco}
\end{table}

\paragraph{Comparison to ResNe(X)t} Table~\ref{exp:coco}(a) lists the results of Swin-T and ResNet-50 on the four object detection frameworks. Our Swin-T architecture brings consistent +3.4$\sim$4.2 box AP gains over ResNet-50, with slightly larger model size, FLOPs and latency. 

Table~\ref{exp:coco}(b) compares Swin Transformer and ResNe(X)t under different model capacity using Cascade Mask R-CNN. Swin Transformer achieves a high detection accuracy of 51.9 box AP and 45.0 mask AP, which are significant gains of +3.6 box AP and +3.3 mask AP over ResNeXt101-64x4d, which has similar model size, FLOPs and latency. On a higher baseline of 52.3 box AP and 46.0 mask AP using an improved HTC framework, the gains by Swin Transformer are also high, at +4.1 box AP and +3.1 mask AP (see Table~\ref{exp:coco}(c)). Regarding inference speed, while ResNe(X)t is built by highly optimized Cudnn functions, our architecture is implemented with built-in PyTorch functions that are not all well-optimized. A thorough kernel optimization is beyond the scope of this paper.

\paragraph{Comparison to DeiT} The performance of DeiT-S using the Cascade Mask R-CNN framework is shown in Table~\ref{exp:coco}(b). The results of Swin-T are +2.5 box AP and +2.3 mask AP higher than DeiT-S with similar model size (86M vs. 80M) and significantly higher inference speed (15.3 FPS vs. 10.4 FPS). The lower inference speed of DeiT is mainly due to its quadratic complexity to input image size.

\paragraph{Comparison to previous state-of-the-art} Table~\ref{exp:coco}(c) compares our best results with those of previous state-of-the-art models. Our best model achieves 58.7 box AP and 51.1 mask AP on COCO test-dev, surpassing the previous best results by +2.7 box AP (Copy-paste~\cite{ghiasi2020copy} without external data) and +2.6 mask AP (DetectoRS~\cite{qiao2020detectors}). 

\begin{table}[t]
    \centering
    \small
\addtolength{\tabcolsep}{-4.1pt}
\begin{tabular}{cc|cc|ccc}
\Xhline{1.0pt}
\multicolumn{2}{c|}{ADE20K}& val & test & \multirow{2}{*}{\#param.} & \multirow{2}{*}{FLOPs} & \multirow{2}{*}{FPS} \\
Method & Backbone & mIoU & score &  &  &  \\
\hline
DANet~\cite{fu2019danet} & ResNet-101 & 45.2 & - & 69M & 1119G & 15.2 \\
DLab.v3+~\cite{chen2018dlabv3p} & ResNet-101 & 44.1 & - & 63M & 1021G & 16.0 \\
ACNet~\cite{fu2019acnet} & ResNet-101 & 45.9 & 38.5 & - & & \\
DNL~\cite{yin2020DNL} & ResNet-101 & 46.0 & 56.2 & 69M & 1249G & 14.8 \\
OCRNet~\cite{yuan2019ocr} & ResNet-101 & 45.3 & 56.0 & 56M & 923G & 19.3 \\
UperNet~\cite{xiao2018upernet} & ResNet-101 & 44.9 & - & 86M & 1029G & 20.1 \\
\hline
OCRNet~\cite{yuan2019ocr} & HRNet-w48 & 45.7 & - & 71M & 664G & 12.5 \\
DLab.v3+~\cite{chen2018dlabv3p} & ResNeSt-101 & 46.9 & 55.1 & 66M & 1051G & 11.9 \\
DLab.v3+~\cite{chen2018dlabv3p} & ResNeSt-200 & 48.4 & - & 88M & 1381G & 8.1 \\
SETR~\cite{zheng2020SETR} & T-Large$^\ddag$ & 50.3 & 61.7 & 308M & - & - \\
\hline
UperNet & DeiT-S$^\dag$ & 44.0 & - & 52M & 1099G & 16.2 \\
\hline
UperNet & Swin-T & 46.1 & - & 60M & 945G & 18.5 \\
UperNet & Swin-S & 49.3 & - & 81M & 1038G & 15.2 \\
UperNet & Swin-B$^\ddag$ & {51.6} & - & 121M & 1841G & 8.7 \\
UperNet & Swin-L$^\ddag$ & \textbf{53.5} & \textbf{62.8} & 234M & 3230G & 6.2 \\
\Xhline{1.0pt}
\end{tabular}
    \caption{Results of semantic segmentation on the ADE20K val and test set. $^\dag$ indicates additional deconvolution layers are used to produce hierarchical feature maps.  $\ddag$ indicates that the model is pre-trained on ImageNet-22K.}
    \label{tab:seg}
    \normalsize
\end{table}

\subsection{Semantic Segmentation on ADE20K}
\paragraph{Settings} ADE20K~\cite{zhou2018semantic} is a widely-used semantic segmentation dataset, covering a broad range of 150 semantic categories. It has 25K images in total, with 20K for training, 2K for validation, and another 3K for testing. We utilize UperNet~\cite{xiao2018upernet} in mmseg~\cite{mmseg2020} as our base framework for its high efficiency.  More details are presented in the Appendix.

\paragraph{Results} Table~\ref{tab:seg} lists the mIoU, model size (\#param), FLOPs and FPS for different method/backbone pairs. From these results, it can be seen that Swin-S is +5.3 mIoU higher (49.3 vs. 44.0) than DeiT-S with similar computation cost. It is also +4.4 mIoU higher than ResNet-101, and +2.4 mIoU higher than ResNeSt-101~\cite{zhang2020resnest}. Our Swin-L model with ImageNet-22K pre-training achieves 53.5 mIoU on the val set, surpassing the previous best model by +3.2 mIoU (50.3 mIoU by SETR~\cite{zheng2020SETR} which has a larger model size).

\begin{table}[]
\small
\centering
\addtolength{\tabcolsep}{-2.5pt}
\begin{tabular}{c|cc|cc|c}
\Xhline{1.0pt}
 & \multicolumn{2}{c|}{ImageNet} & \multicolumn{2}{c|}{COCO} & \multicolumn{1}{c}{ADE20k} \\
 & top-1 & top-5  & AP$^\text{box}$ & AP$^\text{mask}$ & mIoU \\
\hline
w/o shifting & 80.2 & 95.1 & 47.7 & 41.5 & 43.3 \\
shifted windows & \textbf{81.3} & \textbf{95.6} & \textbf{50.5} & \textbf{43.7} & \textbf{46.1} \\
\hline
no pos. & 80.1 & 94.9 & 49.2 & 42.6  & 43.8 \\
abs. pos. & 80.5 & 95.2 & 49.0 & 42.4  & 43.2 \\
abs.+rel. pos. & 81.3 & 95.6 & 50.2 & 43.4 & 44.0\\
rel. pos. w/o app. & 79.3 & 94.7 & 48.2 & 41.9 & 44.1 \\
rel. pos. & \textbf{81.3} & \textbf{95.6} & \textbf{50.5} & \textbf{43.7} & \textbf{46.1} \\
\Xhline{1.0pt}
\end{tabular}
\caption{Ablation study on the \emph{shifted windows} approach and different position embedding methods on three benchmarks, using the Swin-T architecture. w/o shifting: all self-attention modules adopt regular window partitioning, without \emph{shifting}; abs. pos.: absolute position embedding term of ViT; rel. pos.: the default settings with an additional relative position bias term (see Eq.~(\ref{eq.att})); app.: the first scaled dot-product term in Eq.~(\ref{eq.att}).  }
\label{exp:ablation}
\normalsize
\end{table}

\subsection{Ablation Study}

In this section, we ablate important design elements in the proposed Swin Transformer, using ImageNet-1K image classification, Cascade Mask R-CNN on COCO object detection, and UperNet on ADE20K semantic segmentation.

\paragraph{Shifted windows}
Ablations of the \emph{shifted window} approach on the three tasks are reported in Table~\ref{exp:ablation}. Swin-T with the shifted window partitioning outperforms the counterpart built on a single window partitioning at each stage by +1.1\% top-1 accuracy on ImageNet-1K, +2.8 box AP/+2.2 mask AP on COCO, and +2.8 mIoU on ADE20K. The results indicate the effectiveness of using shifted windows to build connections among windows in the preceding layers. The latency overhead by \emph{shifted window} is also small, as shown in Table~\ref{tab:ablation-selfatt-efficient}.

\paragraph{Relative position bias} Table~\ref{exp:ablation} shows comparisons of different position embedding approaches. Swin-T with relative position bias yields +1.2\%/+0.8\% top-1 accuracy on ImageNet-1K, +1.3/+1.5 box AP and +1.1/+1.3 mask AP on COCO, and +2.3/+2.9 mIoU on ADE20K in relation to those without position encoding and with absolute position embedding, respectively, indicating the effectiveness of the relative position bias. Also note that while the inclusion of absolute position embedding improves image classification accuracy (+0.4\%), it harms object detection and semantic segmentation (-0.2 box/mask AP on COCO and -0.6 mIoU on ADE20K). 

While the recent ViT/DeiT models abandon translation invariance in image classification even though it has long been shown to be crucial for visual modeling, we find that inductive bias that encourages certain translation invariance is still preferable for general-purpose visual modeling, particularly for the dense prediction tasks of object detection and semantic segmentation.

\paragraph{Different self-attention methods} The real speed of different self-attention computation methods and implementations are compared in Table~\ref{tab:ablation-selfatt-efficient}. Our cyclic implementation is more hardware efficient than naive padding, particularly for deeper stages. Overall, it brings a 13\%, 18\% and 18\% speed-up on Swin-T, Swin-S and Swin-B, respectively.

The self-attention modules built on the proposed \emph{shifted window} approach are 40.8$\times$/2.5$\times$, 20.2$\times$/2.5$\times$, 9.3$\times$/2.1$\times$, and 7.6$\times$/1.8$\times$ more efficient than those of \emph{sliding windows} in naive/kernel implementations on four network stages, respectively. Overall, the Swin Transformer architectures built on \emph{shifted windows} are 4.1/1.5, 4.0/1.5, 3.6/1.5 times faster than variants built on \emph{sliding windows} for Swin-T, Swin-S, and Swin-B, respectively. Table~\ref{exp:abaltion-selfatt-acc} compares their accuracy on the three tasks, showing that they are similarly accurate in visual modeling.

Compared to Performer~\cite{choromanski2020performer}, which is one of the fastest Transformer architectures (see~\cite{tay2020long}), the proposed \emph{shifted window} based self-attention computation and the overall Swin Transformer architectures are slightly faster (see Table~\ref{tab:ablation-selfatt-efficient}), while achieving +2.3\% top-1 accuracy compared to Performer on ImageNet-1K using Swin-T (see Table~\ref{exp:abaltion-selfatt-acc}).

\begin{table}[]
    \centering
\small
\addtolength{\tabcolsep}{-4.0pt}
\begin{tabular}{c|cccc|ccc}
\Xhline{1.0pt}
\multirow{2}{*}{method} & \multicolumn{4}{c|}{MSA in a stage (ms)} & \multicolumn{3}{c}{Arch. (FPS)} \\
& S1 & S2 & S3 & S4 & T & S & B \\
\hline
sliding window (naive) & 122.5 & 38.3 & 12.1 & 7.6 & 183 & 109 & 77 \\
sliding window (kernel)  & 7.6 & 4.7 & 2.7 & 1.8 & 488 & 283 & 187 \\
\hline
Performer~\cite{choromanski2020performer} & 4.8 & 2.8 & 1.8 & 1.5 & 638 & 370 & 241 \\
\hline
window (w/o shifting) & 2.8 & 1.7 & 1.2 & 0.9 & 770 & 444 & 280 \\
\hline
shifted window (padding) & 3.3 & 2.3 & 1.9 & 2.2 & 670 & 371 & 236 \\
shifted window (cyclic)  & 3.0 & 1.9 & 1.3 & 1.0 & 755 & 437 & 278 \\
\Xhline{1.0pt}
\end{tabular}
    \caption{Real speed of different self-attention computation methods and implementations on a V100 GPU. }
    \label{tab:ablation-selfatt-efficient}
\normalsize
\end{table}

\begin{table}[]
\small
\centering
\addtolength{\tabcolsep}{-4.5pt}
\begin{tabular}{c|c|cc|cc|c}
\Xhline{1.0pt}
& & \multicolumn{2}{c|}{ImageNet} & \multicolumn{2}{c|}{COCO} & \multicolumn{1}{c}{ADE20k} \\
& Backbone & top-1 & top-5  & AP$^\text{box}$ & AP$^\text{mask}$ & mIoU \\
\hline
sliding window & Swin-T & 81.4 & 95.6& 50.2 & 43.5 & 45.8 \\
Performer~\cite{choromanski2020performer} & Swin-T & 79.0 & 94.2 & - & - & -\\
\hline
shifted window & Swin-T & 81.3 & 95.6 & 50.5 & 43.7 & 46.1 \\
\Xhline{1.0pt}
\end{tabular}
\caption{Accuracy of Swin Transformer using different methods for self-attention computation on three benchmarks.}
\label{exp:abaltion-selfatt-acc}
\normalsize
\vspace{-.5em}
\end{table}

\section{Conclusion}

This paper presents Swin Transformer, a new vision Transformer which produces a hierarchical feature representation and has linear computational complexity with respect to input image size. Swin Transformer achieves the state-of-the-art performance on COCO object detection and ADE20K semantic segmentation, significantly surpassing previous best methods. We hope that Swin Transformer’s strong performance on various vision problems will encourage unified modeling of vision and language signals.

As a key element of Swin Transformer, the \emph{shifted window} based self-attention is shown to be effective and efficient on vision problems, and we look forward to investigating its use in natural language processing as well.

\section*{Acknowledgement}

We thank many colleagues at Microsoft for their help, in particular, Li Dong and Furu Wei for useful discussions; Bin Xiao, Lu Yuan and Lei Zhang for help on datasets.

\appendix

\begin{table*}[t]
\small
\centering
\addtolength{\tabcolsep}{-2pt}
\begin{tabular}{c|c|c|c|c|c}
 & \begin{tabular}[c]{@{}c@{}}downsp. rate \\ (output size)\end{tabular} & Swin-T  & Swin-S & Swin-B & Swin-L \\
\hline
\hline
\multirow{3}{*}{stage 1} & \multirow{3}{*}{\begin{tabular}[c]{@{}c@{}}4$\times$\\ (56$\times$56)\end{tabular}} & concat 4$\times$4, 96-d, LN  & concat 4$\times$4, 96-d, LN  & concat 4$\times$4, 128-d, LN   & concat 4$\times$4, 192-d, LN  \\
\cline{3-6}
& & $\begin{bmatrix}\text{win. sz. 7$\times$7,}\\\text{dim 96, head 3}\end{bmatrix}$ $\times$ 2   & $\begin{bmatrix}\text{win. sz. 7$\times$7,}\\\text{dim 96, head 3}\end{bmatrix}$ $\times$ 2    & $\begin{bmatrix}\text{win. sz. 7$\times$7,}\\\text{dim 128, head 4}\end{bmatrix}$ $\times$ 2   & $\begin{bmatrix}\text{win. sz. 7$\times$7,}\\\text{dim 192, head 6}\end{bmatrix}$ $\times$ 2   \\
\hline
\multirow{3}{*}{stage 2}  & \multirow{3}{*}{\begin{tabular}[c]{@{}c@{}}8$\times$\\ (28$\times$28)\end{tabular}} & concat 2$\times$2, 192-d , LN & concat 2$\times$2, 192-d , LN & concat 2$\times$2, 256-d , LN & concat 2$\times$2, 384-d , LN \\
\cline{3-6}
& & $\begin{bmatrix}\text{win. sz. 7$\times$7,}\\\text{dim 192, head 6}\end{bmatrix}$ $\times$ 2  & $\begin{bmatrix}\text{win. sz. 7$\times$7,}\\\text{dim 192, head 6}\end{bmatrix}$ $\times$ 2 & $\begin{bmatrix}\text{win. sz. 7$\times$7,}\\\text{dim 256, head 8}\end{bmatrix}$ $\times$ 2 & $\begin{bmatrix}\text{win. sz. 7$\times$7,}\\\text{dim 384, head 12}\end{bmatrix}$ $\times$ 2 \\
\hline
\multirow{3}{*}{stage 3}  & \multirow{3}{*}{\begin{tabular}[c]{@{}c@{}}16$\times$\\ (14$\times$14)\end{tabular}}  & concat 2$\times$2, 384-d , LN & concat 2$\times$2, 384-d , LN & concat 2$\times$2, 512-d , LN & concat 2$\times$2, 768-d , LN \\
\cline{3-6}
& & $\begin{bmatrix}\text{win. sz. 7$\times$7,}\\\text{dim 384, head 12}\end{bmatrix}$ $\times$ 6 & $\begin{bmatrix}\text{win. sz. 7$\times$7,}\\\text{dim 384, head 12}\end{bmatrix}$ $\times$ 18 & $\begin{bmatrix}\text{win. sz. 7$\times$7,}\\\text{dim 512, head 16}\end{bmatrix}$ $\times$ 18  & $\begin{bmatrix}\text{win. sz. 7$\times$7,}\\\text{dim 768, head 24}\end{bmatrix}$ $\times$ 18 \\
\hline
\multirow{3}{*}{stage 4} & \multirow{3}{*}{\begin{tabular}[c]{@{}c@{}}32$\times$\\ (7$\times$7)\end{tabular}}  & concat 2$\times$2, 768-d , LN & concat 2$\times$2, 768-d , LN & concat 2$\times$2, 1024-d , LN  & concat 2$\times$2, 1536-d , LN \\
\cline{3-6}
& & $\begin{bmatrix}\text{win. sz. 7$\times$7,}\\\text{dim 768, head 24}\end{bmatrix}$ $\times$ 2 & $\begin{bmatrix}\text{win. sz. 7$\times$7,}\\\text{dim 768, head 24}\end{bmatrix}$ $\times$ 2  & $\begin{bmatrix}\text{win. sz. 7$\times$7,}\\\text{dim 1024, head 32}\end{bmatrix}$ $\times$ 2 & $\begin{bmatrix}\text{win. sz. 7$\times$7,}\\\text{dim 1536, head 48}\end{bmatrix}$ $\times$ 2 \\
\end{tabular}
\normalsize
\caption{Detailed architecture specifications. }
\label{table:arch-spec}
\end{table*}

\renewcommand{\thesection}{A\arabic{section}}

\section{Detailed Architectures}

The detailed architecture specifications are shown in Table~\ref{table:arch-spec}, where an input image size of 224$\times$224 is assumed for all architectures.
``Concat $n\times n$'' indicates a concatenation of $n\times n$ neighboring features in a patch. This operation results in a downsampling of the feature map by a rate of $n$. ``96-d'' denotes a linear layer with an output dimension of 96. ``win. sz. $7\times7$'' indicates a multi-head self-attention module with window size of $7\times 7$.

\section{Detailed Experimental Settings}

\subsection{Image classification on ImageNet-1K}

The image classification is performed by applying a global average pooling layer on the output feature map of the last stage, followed by a linear classifier. We find this strategy to be as accurate as using an additional \texttt{class} token as in ViT~\cite{dosovitskiy2020vit} and DeiT~\cite{touvron2020deit}. In evaluation, the top-1 accuracy using a single crop is reported.

\paragraph{Regular ImageNet-1K training} The training settings mostly follow~\cite{touvron2020deit}. For all model variants, we adopt a default input image resolution of 224$^2$. For other resolutions such as 384$^2$, we fine-tune the models trained at 224$^2$ resolution, instead of training from scratch, to reduce GPU consumption.

When training from scratch with a 224$^2$ input, we employ an AdamW~\cite{kingma2014adam} optimizer for 300 epochs using a cosine decay learning rate scheduler with 20 epochs of linear warm-up.  A batch size of 1024, an initial learning rate of 0.001, a weight decay of 0.05, and gradient clipping with a max norm of 1 are used. We include most of the augmentation and regularization strategies of~\cite{touvron2020deit} in training, including RandAugment~\cite{cubuk2020randaugment}, Mixup~\cite{zhang2017mixup}, Cutmix~\cite{yun2019cutmix}, random erasing~\cite{zhong2020random} and stochastic depth~\cite{huang2016deep}, but not repeated augmentation~\cite{hoffer2020augment} and Exponential Moving Average (EMA)~\cite{polyak1992acceleration} which do not enhance performance. Note that this is contrary to~\cite{touvron2020deit} where repeated augmentation is crucial to stabilize the training of ViT. An increasing degree of stochastic depth augmentation is employed for larger models, i.e. $0.2, 0.3, 0.5$ for Swin-T, Swin-S, and Swin-B, respectively.

For fine-tuning on input with larger resolution, we employ an adamW~\cite{kingma2014adam} optimizer for 30 epochs with a constant learning rate of $10^{-5}$, weight decay of $10^{-8}$, and the same data augmentation and regularizations as the first stage except for setting the stochastic depth ratio to 0.1.

\paragraph{ImageNet-22K pre-training} We also pre-train on the larger ImageNet-22K dataset, which contains 14.2 million images and 22K classes. The training is done in two stages. For the first stage with 224$^2$ input, we employ an AdamW optimizer for 90 epochs using a linear decay learning rate scheduler with a 5-epoch linear warm-up. A batch size of 4096, an initial learning rate of 0.001, and a weight decay of 0.01 are used. In the second stage of ImageNet-1K fine-tuning with 224$^2$/384$^2$ input, we train the models for 30 epochs with a batch size of 1024, a constant learning rate of $10^{-5}$, and a weight decay of $10^{-8}$.

\subsection{Object detection on COCO}

For an ablation study, we consider four typical object detection frameworks: Cascade Mask R-CNN~\cite{he2017mask,cai2018cascade}, ATSS~\cite{zhang2020bridging}, RepPoints v2~\cite{chen2020reppointsv2}, and Sparse RCNN~\cite{sun2020sparsercnn} in mmdetection~\cite{chen2019mmdetection}. For these four frameworks, we utilize the same settings: multi-scale training~\cite{carion2020detr,sun2020sparsercnn} (resizing the input such that the shorter side is between 480 and 800 while the longer side is at most 1333), AdamW~\cite{loshchilov2017decoupled} optimizer (initial learning rate of 0.0001, weight decay of 0.05, and batch size of 16), and 3x schedule (36 epochs with the learning rate decayed by $10\times$ at epochs 27 and 33).

For system-level comparison, we adopt an improved HTC~\cite{chen2019htc} (denoted as HTC++) with instaboost~\cite{fang2019instaboost}, stronger multi-scale training~\cite{cao2019gcnet} (resizing the input such that the shorter side is between 400 and 1400 while the longer side is at most 1600), 6x schedule (72 epochs with the learning rate decayed at epochs 63 and 69 by a factor of 0.1), soft-NMS~\cite{Bodla2017softnms}, and an extra global self-attention layer appended at the output of last stage and ImageNet-22K pre-trained model as initialization. We adopt stochastic depth with ratio of $0.2$ for all Swin Transformer models.

\subsection{Semantic segmentation on ADE20K}

ADE20K~\cite{zhou2018semantic} is a widely-used semantic segmentation dataset, covering a broad range of 150 semantic categories. It has 25K images in total, with 20K for training, 2K for validation, and another 3K for testing. We utilize UperNet~\cite{xiao2018upernet} in mmsegmentation~\cite{mmseg2020} as our base framework for its high efficiency. 

In training, we employ the AdamW~\cite{loshchilov2017decoupled} optimizer with an initial learning rate of $6\times10^{-5}$, a weight decay of 0.01, a scheduler that uses linear learning rate decay, and a linear warmup of 1,500 iterations. Models are trained on 8 GPUs with 2 images per GPU for 160K iterations. For augmentations, we adopt the default setting in mmsegmentation of random horizontal flipping, random re-scaling within ratio range [0.5, 2.0] and random photometric distortion. Stochastic depth with ratio of $0.2$ is applied for all Swin Transformer models.
Swin-T, Swin-S are trained on the standard setting as the previous approaches with an input of 512$\times$512. Swin-B and Swin-L with $\ddag$ indicate that these two models are pre-trained on ImageNet-22K, and trained with the input of 640$\times$640.

In inference, a multi-scale test using resolutions that are [0.5, 0.75, 1.0, 1.25, 1.5, 1.75]$\times$ of that in training is employed. When reporting test scores, both the training images and validation images are used for training, following common practice~\cite{yin2020DNL}.

\begin{table}[]
    \centering
    \small
\addtolength{\tabcolsep}{-3pt}
    \begin{tabular}{c|cc|cc|cc}
\Xhline{1.0pt}
& \multicolumn{2}{c|}{Swin-T}& \multicolumn{2}{c|}{Swin-S}& \multicolumn{2}{c}{Swin-B} \\
\begin{tabular}[c]{@{}c@{}}input\\ size\end{tabular} & \begin{tabular}[c]{@{}c@{}}top-1\\ acc\end{tabular} & \begin{tabular}[c]{@{}c@{}}throughput\\ (image / s)\end{tabular} & \begin{tabular}[c]{@{}c@{}}top-1\\ acc\end{tabular} & \begin{tabular}[c]{@{}c@{}}throughput\\ (image / s)\end{tabular} & \begin{tabular}[c]{@{}c@{}}top-1\\ acc\end{tabular} & \begin{tabular}[c]{@{}c@{}}throughput\\ (image / s)\end{tabular} \\
\hline
224$^2$ & 81.3 & 755.2 & 83.0 & 436.9 & 83.3 & 278.1 \\
256$^2$ & 81.6 & 580.9 & 83.4 & 336.7 & 83.7 & 208.1 \\
320$^2$ & 82.1 & 342.0 & 83.7 & 198.2 & 84.0 & 132.0 \\
384$^2$ & 82.2 & 219.5 & 83.9 & 127.6 & 84.5 & 84.7 \\
\Xhline{1.0pt}
\end{tabular}
    \caption{Swin Transformers with different input image size on ImageNet-1K classification.}
    \label{tab:diff-input}
    \normalsize
\end{table}

\section{More Experiments}

\subsection{Image classification with different input size}

Table~\ref{tab:diff-input} lists the performance of Swin Transformers with different input image sizes from $224^2$ to $384^2$. In general, a larger input resolution leads to better top-1 accuracy but with slower inference speed.

\begin{table}[]
    \centering
    \small
\addtolength{\tabcolsep}{-4.8pt}
\begin{tabular}{cc|ccc|ccc}
\Xhline{1.0pt}
Backbone& Optimizer & AP$^\text{box}$ & AP$^\text{box}_\text{50}$ & AP$^\text{box}_\text{75}$ & AP$^\text{mask}$ & AP$^\text{mask}_\text{50}$ & AP$^\text{mask}_\text{75}$ \\
\hline
\multirow{2}{*}{R50} & SGD & 45.0 & 62.9 & 48.8 & 38.5 & 59.9 & 41.4 \\
& AdamW & 46.3 & 64.3 & 50.5 & 40.1 & 61.7 & 43.4 \\
\hline
\multirow{2}{*}{X101-32x4d} & SGD & 47.8 & 65.9 & 51.9 & 40.4 & 62.9 & 43.5 \\
& AdamW & 48.1 & 66.5 & 52.4 & 41.6 & 63.9 & 45.2 \\
\hline
\multirow{2}{*}{X101-64x4d} & SGD & 48.8 & 66.9 & 53.0 & 41.4 & 63.9 & 44.7 \\
& AdamW & 48.3 & 66.4 & 52.3 & 41.7 & 64.0 & 45.1 \\
\Xhline{1.0pt}
\end{tabular}
    \caption{Comparison of the SGD and AdamW optimizers for ResNe(X)t backbones on COCO object detection using the Cascade Mask R-CNN framework.}
    \label{tab:app-optimizer}
    \normalsize
\end{table}

\subsection{Different Optimizers for ResNe(X)t on COCO} 

Table~\ref{tab:app-optimizer} compares the AdamW and SGD optimizers of the ResNe(X)t backbones on COCO object detection. The Cascade Mask R-CNN framework is used in this comparison. While SGD is used as a default optimizer for Cascade Mask R-CNN framework, we generally observe improved accuracy by replacing it with an AdamW optimizer, particularly for smaller backbones. We thus use AdamW for ResNe(X)t backbones when compared to the proposed Swin Transformer architectures.

\subsection{Swin MLP-Mixer}

We apply the proposed hierarchical design and the shifted window approach to the MLP-Mixer architectures~\cite{tolstikhin2021mlpmixer}, referred to as Swin-Mixer. Table~\ref{exp:imagenet-swin-mlp-mixer} shows the performance of Swin-Mixer compared to the original MLP-Mixer architectures MLP-Mixer~\cite{tolstikhin2021mlpmixer} and a follow-up approach, ResMLP~\cite{tolstikhin2021mlpmixer}. Swin-Mixer performs significantly better than MLP-Mixer (81.3\% vs. 76.4\%) using slightly smaller computation budget (10.4G vs. 12.7G). It also has better speed accuracy trade-off compared to ResMLP~\cite{touvron2021resmlp}. These results indicate the proposed hierarchical design and the shifted window approach are generalizable.

\begin{table}[t]
    \centering
    \small
    \addtolength{\tabcolsep}{-5.pt}
    \begin{tabular}{c|cccc|c}
    \Xhline{1.0pt}
    method &  \begin{tabular}[c]{@{}c@{}}image \\ size\end{tabular} & \#param. & FLOPs & \begin{tabular}[c]{@{}c@{}}throughput\\ (image / s)\end{tabular} & \begin{tabular}[c]{@{}c@{}}ImageNet \\ top-1 acc.\end{tabular} \\
    \hline
    MLP-Mixer-B/16~\cite{tolstikhin2021mlpmixer} & 224$^2$ & 59M & 12.7G & - & 76.4 \\
    ResMLP-S24~\cite{touvron2021resmlp} & 224$^2$ & 30M & 6.0G & 715 & 79.4 \\
    ResMLP-B24~\cite{touvron2021resmlp} & 224$^2$ & 116M & 23.0G & 231 & 81.0 \\
    \hline
    \makecell{Swin-T/D24\\(Transformer)} & 256$^2$ & 28M & 5.9G & 563 & 81.6 \\
    \hline
    Swin-Mixer-T/D24 & 256$^2$ & 20M & 4.0G & 807 & 79.4 \\
    Swin-Mixer-T/D12 & 256$^2$ & 21M & 4.0G & 792 & 79.6 \\
    Swin-Mixer-T/D6 & 256$^2$ & 23M & 4.0G & 766 & 79.7 \\
    \hline
    {\makecell{Swin-Mixer-B/D24\\(no shift)}} & 224$^2$ & 61M & 10.4G & 409 & 80.3 \\
    Swin-Mixer-B/D24 & 224$^2$ & 61M & 10.4G & 409 & 81.3 \\
    \Xhline{1.0pt}
    \end{tabular}
    \normalsize
    \caption{Performance of Swin MLP-Mixer on ImageNet-1K classification. $D$ indictes the number of channels per head. Throughput is measured using the GitHub repository of~\cite{rw2019timm} and a V100 GPU, following~\cite{touvron2020deit}.}
    \label{exp:imagenet-swin-mlp-mixer}
    \end{table}

{\small
\bibliographystyle{ieee_fullname}
\bibliography{egbib}
}

\end{document}